\documentclass{egpubl}
\usepackage{eg2024}

\CGFStandardLicense
\usepackage[T1]{fontenc}
\usepackage{dfadobe}  

\usepackage{cite}  % comment out for biblatex with backend=biber
\BibtexOrBiblatex
\electronicVersion
\PrintedOrElectronic
\ifpdf \usepackage[pdftex]{graphicx} \pdfcompresslevel=9
\else \usepackage[dvips]{graphicx} \fi

\usepackage{egweblnk} 

\usepackage[table]{xcolor}
\usepackage{amssymb}
\usepackage{url}
\usepackage{graphicx}
\usepackage{amsmath}
\usepackage{nicefrac}
\usepackage{booktabs}
\usepackage{multirow}
\usepackage{graphicx}
\usepackage[colorinlistoftodos]{todonotes}
\definecolor{rvcolor}{RGB}{0, 0, 0}

\newcommand{\RV}[1]{\textcolor{rvcolor}{#1}}

\DeclareMathOperator*{\argmin}{arg\,min}

\definecolor{cellcolor}{HTML}{EFEFEF}

\newcommand{\NBC}{BCf-}
\newcommand{\unq}[1]{\overline{#1}} % we also can use \widehat ?

\newcommand{\MLP}{MLP}
\newcommand{\bppc}{bppc}
\newcommand{\ARM}{arm}

\usepackage{mathtools}
\usepackage{xspace} 
\usepackage{url}

\author[C. Weinreich \& L. De~Oliveira]
{
    {\parbox{\textwidth}{\centering 
        C. Weinreich\thanks{Equal Contribution, order determined by coin toss}, 
        L. De~Oliveira\footnotemark[1], 
        A. Houdard\footnotemark[1] and 
        G. Nader\footnotemark[1]
        }
    }\\
    {\parbox{\textwidth}{\centering Ubisoft La Forge}}
}

\title{Real-Time Neural Materials using Block-Compressed Features}
\begin{document}

\teaser{
 \centering
 \includegraphics[width=0.8\linewidth]{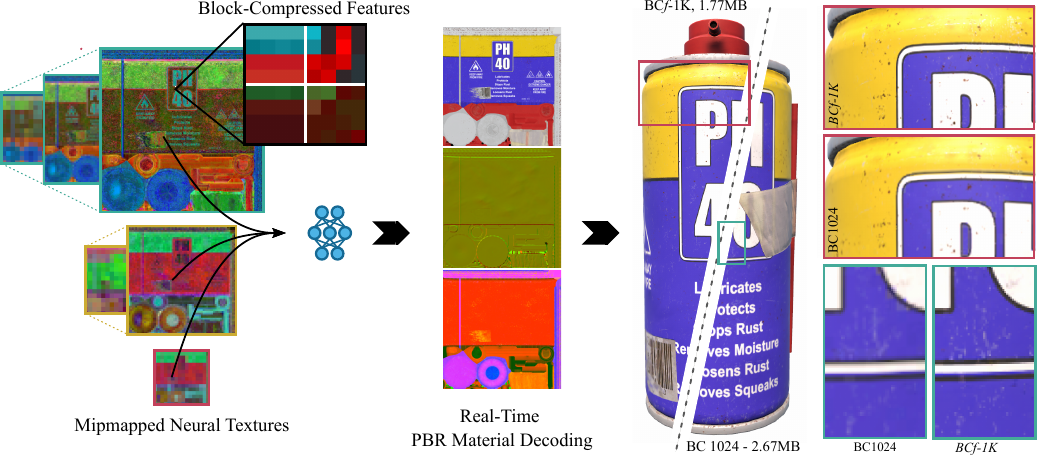}
 \caption{PBR material texture set is decoded in real-time from our Block Compressed neural features (BCf) resulting in an image that is visually sharper than standard BC textures of similar resolution.}
 \label{fig:teaser}
}

\maketitle

\begin{abstract}

% probleme
Neural materials typically consist of a collection of neural features along with a decoder network.
The main challenge in integrating such models in real-time rendering pipelines lies in the large size required to store their features in GPU memory and the complexity of evaluating the network efficiently. 
% solution
We present a neural material model whose features and decoder are specifically designed to be used in real-time rendering pipelines.
Our framework leverages hardware-based block compression (BC) texture formats to store the learned features and trains the model to output the material information continuously in space and scale.
% comment
To achieve this, we organize the features in a block-based manner and emulate BC6 decompression during training, making it possible to export them as regular BC6 textures.
This structure allows us to use high resolution features while maintaining a low memory footprint.
Consequently, this enhances our model's overall capability,
enabling the use of a lightweight and simple decoder architecture that can be evaluated directly in a shader.
Furthermore, since the learned features can be decoded continuously, it allows for random $uv$ sampling and smooth transition between scales without needing any subsequent filtering. 
% resultats
As a result, our neural material has a small memory footprint, can be decoded extremely fast adding a minimal computational overhead to the rendering pipeline. 

\end{abstract}
%==============================================================%
\section{Introduction}

The continuous challenge of real-time rendering systems is to improve the graphics quality while reducing both the memory cost and the evaluation time. Recent progress in graphics hardware and rendering algorithms has gotten us closer to this goal.
For instance, hardware accelerated ray-tracing \cite{rtptcourse2022} facilitates real-time dynamic global illumination and drastically improves the quality of shadows, reflections  and refractions \cite{lumencourse2022, ouyang2021restir, datta2022neural}.
Besides that, micropolygon-based data structures \cite{maggiordomo2023micro,nanitecourse2021} allows the rendering of extremely detailed and dense scenes.
In terms of visual aesthetics, the Physically Based Rendering (PBR) framework has been the standard in real-time applications for well over a decade \cite{PBRCourse2012, PBRCourse2013, hill2020physically}.
In this framework, a material is composed of multiple texture layers (such as albedo, normals, metalness, etc.) with each layer serving a distinct role in accurately representing various aspects of the material's visual characteristics. 
Enhancing the visual quality of materials typically involves the process of either layering multiple elements \cite{matcourse20223} or increasing the texture resolution, which affects their computational cost and memory footprint, respectively. 
Consequently, rendering hyperrealistic materials in real-time applications, such as video games, remains somewhat restricted. 

Recent work has shown the potential of neural networks to model and represent material properties \cite{sztrajman2021neural, zheng2021compact, Fan2022NLBRDF, Xu2023NeuSample}. 
This neural approach aims at replacing traditional PBR textures with a collection of learned latent features, also known as neural textures \cite{thies2019deferred}, alongside a neural network, usually a Multi-Layer Perceptron (MLP).
In this context, the network plays a crucial role in decoding the learned information and reconstructing the original material. 
\RV{
While neural approaches have proven successful in accelerating the rendering of complex appearances \cite{zeltner2023real} and compressing high-resolution PBR materials \cite{ntc2023}, its integration in consumer oriented real-time applications such as video games is not straightforward.
}
% There are two primary challenges that contribute to this. 
\RV{
The trained latent features have a relatively big memory footprint as they are often stored in GPU memory using an uncompressed format.   
Recent work by Vaidyanathan \textit{et al.} \cite{ntc2023} proposed to overcome this issue by reducing the resolution and heavily quantize the values of these features.
However, to compensate for the loss in resolution and precision, the associated decoder network is rather large and thus computationally intensive which leads slow evaluation time. 
As a result, achieving real-time performance remains challenging, and depends on specific hardware extensions that are supported on only the most recent high-end hardware.
The method we present addresses these challenges.
Our goal is to design a neural material model that can be integrated into a rendering pipeline and achieve real-time performance without having to rely on any specific hardware acceleration. 
}

To do so, we present a framework capable of
(1) learning the material information at any point and scale and 
(2) leveraging hardware-based Block Compression (BC) format to store the learned neural features.
Our features can thus be decoded continuously at any scale in $uv$ space, making it possible to reconstruct the material information with one sample per pixel and achieve smooth transition between various scales.
This removes the need for any subsequent filtering step.
Furthermore, encoding our neural features as regular block compressed textures reduces their memory footprint enabling us to increase their resolution. 
This, in turn, directly influences the complexity of the decoder network, making it considerably simpler and significantly reduces computational time.

The paper is organized as follows. 
Section \ref{sec:relatedwork} quickly details existing material representation and compression methods. 
We then present the technical aspects of our neural material learning framework and block based features in sections \ref{sec:framework} and \ref{sec:bclayers} respectively.
In section \ref{sec:practical} we describe how to practically use our framework to learn a standard PBR material and integrate it in a real-time rendering pipeline.
Finally, we present our results in section \ref{sec:results} and conclude by discussing some limitations and future work in section \ref{sec:limitations}.

\section{Related Work}
\label{sec:relatedwork}

This section provides an overview of the various methods to represent and store materials that are relevant to our work. 
We first focus on the traditional texture based approaches (sec.~\ref{sec:tex_material}) then highlight the more recent field of neural representation methods (sec.~\ref{sec:neu_mat}).

\subsection{Texture Materials}
\label{sec:tex_material}
The material properties of a three-dimensional object can be thought of as a multi-dimensional signal, which associates every point $(u, v)$ on its two-dimensional surface with the parameter space of its appearance model. 
In a real-time rendering context, this is primarily done via texture mapping \cite{rtr4} where the material properties are discretely stored in a collection of textures with dimensions $h \times w \times c$. 
Here, $h$ and $w$ denote the spatial resolution of the textures, and $c$ represents the total number of channels across all texture layers in the set.
Increasing the visual fidelity of materials primarily relies on either adding more layers \cite{hill2020physically} or increasing their resolution which has led to a significant rise in memory requirements.
To address this issue, textures are stored in GPU memory in a compressed format.
Due to the object's arbitrary position and orientation, material information are sampled from the textures at random locations. 
It is therefore necessary for the texture's compression format to allow for random-access and filtering so that the information can be decompressed at any point in $uv$ space when needed in real-time.
This makes standard image compression techniques, such as JPEG \cite{wallace92jpeg, alakuijala2019jpeg}, as well as newer neural based ones \cite{balle2018variational, cheng2020learned} not suitable for this application as they require unpacking the entire image before being able to access pixel information.

\begin{figure}
    \centering
    \includegraphics[width=0.8\linewidth]{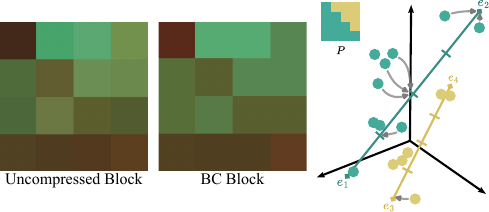}
    \caption{Block Compression algorithms encode a block of $4 \times 4$ pixels with a set of endpoints forming one or multiple line segments and index each pixel based on it's projected position in the RGB space. In the case where two or more line segments are stored, the pixels are separated into groups according to a pre-defined partition $P$.}
    \label{fig:bcformat}
\end{figure}

For real-time rendering, block compression methods \cite{campbell86ccc, bcpatent1} have long been the standard for compressing material textures.
There are seven variants of the BC format (BC1-BC7) supported in DirectX \cite{D3D}, each designed for specific types of image data. 
All BC formats divide the image into block of 4x4 pixels and operate under the assumption that the colors in each block exhibits minimal variation and are evenly distributed along one or more line segments within the RGB color space.
This means that each block can be represented by a very small color palette.
In this context, the information in each block is encoded by storing the two endpoints of each segment and indexing each pixel according to its position in the RGB space (fig. \ref{fig:bcformat}).
Reconstructing the pixel value is simply done by blending the two endpoints proportionally to the index value.
The more recent BC6 and BC7 formats are the ones using more than one segment per block. 
These formats, designed for floating points and RGBA data respectively, improve the compression quality by separating the pixels inside a block into several groups, each having its own set of endpoints.
The groups are chosen from a set of predefined partitions.
In this case, each block stores more than one set of endpoints as well as the partition number.
However, BC formats can only compress textures with up to four channels which is not suited for high dimensional materials. 
To overcome this, the material data is separated into several textures that are compressed independently.
ASTC \cite{nystad2012adaptive} is another popular block-based texture compression technique.
It is more flexible than BC as it supports various block dimensions including non-square blocks and can even handle 3D textures. 

\subsection{Neural Materials}
\label{sec:neu_mat}
Over the last few years, advancement in neural rendering \cite{tewari2022advances} have shown that it is possible to represent a digital signal, such as a material $M$, with a neural network $f_\eta$ by minimizing over its weights $\eta$ the following quantity:
\begin{equation}
   \sum_{i,j} \| f_\eta(u_i,v_j) - M_{i,j} \|^2, 
\end{equation}
where $M_{i,j}$ is the pixel value of the target material and $(u_i, v_j)$ are their corresponding local coordinates.
This is an overfitting problem where the weights of the network are optimized such that the network's output matches the target image at a given pixel. 
Thus, the capability of the network is key to perfectly recover the original image. 
For instance, simple coordinate based networks \cite{mildenhall2020nerf} are not capable of dealing with high frequency details. 
To improve the network's reconstruction capacity, positional encoding \cite{tancik2020fourier, chng2022gaussian} where the input $(u, v)$ coordinates is encoded as a vector generated from a periodic function, is usually employed.
However, this requires a large network to accurately reconstruct the original image which makes it not suitable for real-time evaluation.
Using trainable spatial features \cite{mueller2022instant,Chen2023factorfield,chen2023neurbf,shin2023binary}, \textit{i.e.}, neural textures,
drastically reduces the size of the network and improves the reconstruction quality. 
In this setting, the input coordinates are used to sample in the neural textures using bilinear interpolation and the resulting feature vector is given as input to the neural network.

The idea of pairing a set of discrete spatial features and a neural network have proven to be popular for material representation. 
For instance, Rainer et al. \cite{rainer2019neural} uses an encoder-decoder architecture to compress a large set of Bidirectional Texture Function layers. 
The encoder is trained to generate a latent code for each texel.
This code is then used by the decoder, in conjunction with a light and view direction to output a single RGB value.
Kuznetsov et al. \cite{kuznetsov2021neumip} combined a pyramid of neural textures with a fully connected network to learn the material properties at different scales. 
Their model is capable of rendering materials with intricate parallax effects on an infinite plane but fails to generalize to curved surfaces. 
This was later done in \cite{kuznetsov2022rendering}.
By adding curvature and transparency as part of the network's input and output respectively. 
More recently, Zeltner et al. \cite{zeltner2023real} have used a set of hierarchical textures and two MLPs to bake complex film-quality appearance.
Here, the first network learns the material's reflectance and the seconds produces importance-sampled directions. 
This model is about three times faster than standard node-based multi-layers materials when integrated into a path-tracing pipeline.

Despite this progress, these methods often overlook the issue of storage size. 
In practice, learned neural features are stored in an uncompressed format which can be quite large to practically use in a real-time environment, especially when considering the memory capacity of mainstream hardware.
Vaidyanathan et al \cite{ntc2023} addressed this issue and demonstrated how neural material representations can be more efficient than standard texture compression techniques at storing PBR material information.
To do so, they reduce the resolution of the neural features as much as possible and aggressively quantize their values. 
This made it possible to compress PBR textures at very low bit-rates, up to 0.2 bits per-pixel per-channel (\bppc).
\RV{
However, integrating this neural material decompression in current rendering pipeline and achieving real-time performance requires access to the latest high-end hardware.
}
Moreover, since the model from \cite{ntc2023} only learns the material information at fixed coordinates, it is essential to couple the neural decompression with a filtering operation
\cite{fajardo2023stochastic} in order to minimize flickering and aliasing artefacts.
This introduces additional computational overhead as it entails decompressing multiple samples per pixel.

In the following, we introduce a novel neural material representation using Block-Compressed features (BCf) specifically designed to be integrated into a traditional real-time rendering pipeline at minimal computational overhead.

\begin{figure*}
    \centering
    \includegraphics[width=0.95\textwidth]{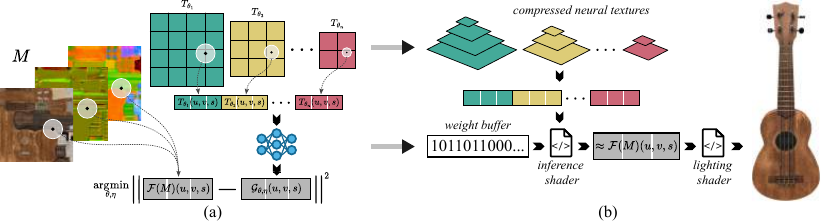}
    \caption{Overview of our neural material framework. (a) The neural features $T_{\theta_i}$ and the MLP $f_\eta$ are fitted through backpropagation to match the filtered material $\mathcal{F}(M)$. (b) After training, the neural features $T_{\theta_i}$ are exported as mipmapped texture sets that can be sampled by the engine and the weights $\eta$ of the MLP are exported as a binary buffer. A shader is used to perform the MLP inference after trilinearly sampling the neural texture, outputting the filtered material $\mathcal{G}_{\theta, \eta}(u, v, s) \approx \mathcal{F}(M)(u,v,s)$. Finally, The renderer can perform the shading step as usual.}
    \label{fig:overview}
\end{figure*}

\section{Neural Material Framework}
\label{sec:framework}

A PBR material $M$ is constituted of a set of properties. These properties are represented at a coordinate $(u, v)$ by a vector of dimension $c$.
In a standard 3D application, this data is stored in a discrete fashion through a texture of size $w \times h \times c$ and mapped onto a 3D object.
At render time, this information is sampled from the textures as follows:
\begin{equation}
    \label{eq:tradmat}
    \mathcal{F}(M)(u,v,s) \in\mathbf{R}^c, 
\end{equation}
where $\mathcal{F}$ is a filtering operation and $s$ is a scale value.
The filtering is essential here.
It accounts for the misalignment and the difference in area between the screen pixel and its corresponding texels, caused by the 3D object's arbitrary distance and orientation.
Our neural material model aims at replacing the material textures with a set of neural features containing abstract information and a decoder network.
The role of the decoder here is to reconstruct the material data at a given point and scale using the learned features.
Figure~\ref{fig:overview} gives an overview of our neural material framework.

Let $ \{ \text{T}_{\theta_0}, \ldots, \text{T}_{\theta_n} \}$ be a set of neural features of size $\left( w_0, h_0, d_0 \right), \ldots, \left( w_n, h_n, d_n \right)$ with trainable parameters $\{ \theta_0 , \ldots , \theta_n \}$. And let $f_\eta$ be a fully connected neural network of input size $\sum_{i=1}^{n}d_i$ and output size $c$ with trainable parameters $\eta$.
To reconstruct the material information at a point $(u, v)$ with respect to a scale $s$, we sample each of the features, concatenate the resulting values 
and pass it to the neural network as follow:
\begin{equation}
    \label{eq:neumat}
    \mathcal{G}_{\theta, \eta}(u, v, s) = f_\eta \left( T_{\theta_0} \left( u, v, s\right), \ldots, T_{\theta_n} \left( u, v, s \right) \right).
\end{equation}
In this framework, learning the material $M$ boils down to optimizing $\theta$ and $\eta$ such that:
\begin{equation}
    \label{eq:minim}
    \hat{\theta}, \hat{\eta} = \argmin_{\theta,\eta} \left\| \mathcal{G}_{\theta, \eta} - \mathcal{F}(M) \right\|^2.
\end{equation}
This allows us to train a model such that it simulates a given filtering operation.
In practice, we use a batched stochastic gradient descent to perform this optimization where the gradient is computed on a batch of random values of $(u, v, s)$. 
This leads us to minimize the following loss:
\begin{equation}
\label{eq:loss}
\sum_{(u,v, s)\in B} \left\| \mathcal{G}_{\theta, \eta}(u,v,s) - \mathcal{F}(M)(u,v,s) \right\|^2.
\end{equation}

\begin{figure}
    \centering
    \includegraphics[width=0.8\linewidth]{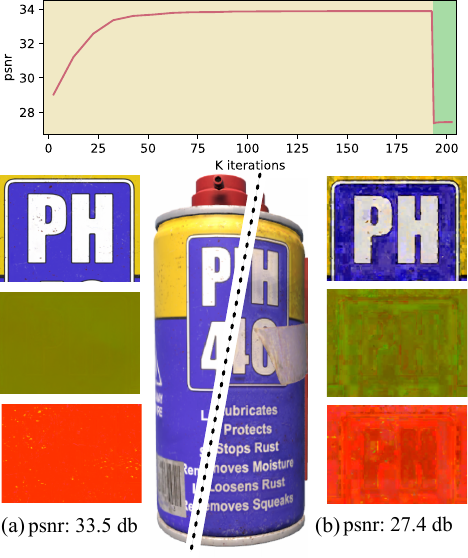}
    \caption{Neural material reconstructed from (a) raw unconstrained neural features and (b) compressed ones. The naive compression of the neural features will lead to artifacts severely affecting the visual quality.}
    \label{fig:naiveneuralbc}
\end{figure}

Once trained, we export the learned features as textures and store the model's weights in a binary buffer. 
At render time, the neural textures and the weights are used to reconstruct the material information. 
To make neural materials more appealing, it is important to reduce their memory footprint and to make their memory size match the traditional material texture.
There are two possible strategies here. 
The first one consists in using low bitrate and low resolution neural features \cite{ntc2023}.
However, this approach leads to a more complex decoder architecture, making the material reconstruction more computationally expensive.
The second aims at storing the neural features more efficiently by using a compression algorithm.
We chose the latter and propose to store them in the BC6 format as it is designed to handle floating point data and allows for random-access.
Additionally, these compressed features can be handled like any traditional texture, which considerably simplifies the neural material's decoding at rendertime.
However, naively compressing the learned features at the end of the training will lead to artifacts that severely impact the visual quality of the reconstructed material (fig. \ref{fig:naiveneuralbc}).
In order to tackle this issue, we propose a specific block compressed neural feature parameterization that is compatible with the BC6 format.

\section{Block Based Neural Features}
\label{sec:bclayers}

In this section, we detail our block-based neural features.
The goal here is to be able to store the trained neural features as BC6 textures without affecting the visual quality and use them in a real-time environment.
To do so, we structure the features in blocks of $4 \times 4$ and define the parameters on a per-block basis (sec. \ref{sec:BCparam}). 
Then, we design a forward pass that emulate the BC6 decompression (sec. \ref{sec:BCforward}) and takes advantages of hardware texture samplers (sec. \ref{sec:bcntfilter}).
This makes it possible to directly export the neural features as BC6 textures without the need for a subsequent compression step and use them into real-time environment with a minimal computational overhead.

\subsection{Block parameterization}
\label{sec:BCparam}

In the BC6 setting, an image is stored by encoding the endpoints value for each $4 \times 4$ block and indexing each pixel according to its distance to the corresponding line segment (fig. \ref{fig:bcformat}).
The dual partition mode of BC6 divides the pixels inside each block into two regions allowing for two lines segments per block.

We design the neural features to mimic this behaviour.
Given a feature layer $T_{\theta_i}$ of size $w \times h \times 3$, we force its values within a block of size $4 \times 4$ to lie on two lines segments.
This is done by design,
we structure our block-based neural features by modeling the parameter set $\theta_i$ such that each block of size $4\times4$ is modeled with
\begin{equation}
    \begin{cases}
        l_1 = \{e_1, e_2\} \subset \mathbf{R}^3\\
        l_2 = \{e_3, e_4\}\subset \mathbf{R}^3\\
        x = \{x_1, \ldots, x_{16}\} \subset [0, 1]
    \end{cases},
\end{equation}
\RV{
where $l_1$ and $l_2$ are two sets of endpoints for the first and second line segments respectively and $x$ pixel index represented as the relative position of each one of the sixteen pixels on the corresponding line segment.
To determine which pixel belongs to which segment, we attach to each block an integer, $k \in \{0, \ldots, 31\}$, linking to a binary mask of the corresponding partition. 
We refer the reader to the DirectX BC documentation \cite{D3D} for all the technical details regarding the pre-defined partitions. 
}
 
\begin{figure}
    \centering
    \includegraphics[width=0.8\linewidth]{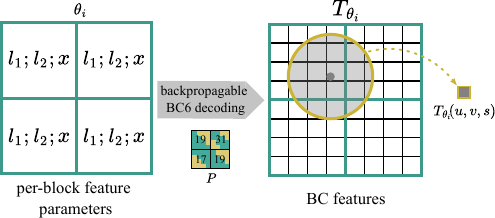}
    \caption{Evaluating $T_{\theta_i}$ at position $(u, v)$ and scale $s$ is done by performing a BC6 decompression then filtering the result with respect to the given scale. Our simulated BC6 decoding makes it possible to backpropagate through this operation and train the per-block neural features.}
    \label{fig:bclayer}
\end{figure}

Reconstructing the material information (eq. \ref{eq:neumat}) requires sampling each neural feature $T_{\theta_i}$ at position $(u, v)$ and scale $s$. 
This consists of performing the BC6 decompression and then filtering the corresponding value according to a certain strategy (fig. \ref{fig:bclayer}).
In order to train our block-based model, it is necessary to backpropagate through this decompression.
Additionally, it is important to choose a filtering strategy that does not incur additional computation overhead.
The latter is crucial to be able to integrate our neural material model in rendering pipelines. 

\subsection{Trainable BC6 decompression}
\label{sec:BCforward}

In the two partition mode, a BC6 block stores two sets of quantized endpoints, the pixel indices and a partition~ID.
The decompression operation uses this data to recover back the original information by mixing the endpoints proportionally with the index values.
According to the BC6 standard specification \cite{D3D}, this operation is non-linear and is done in three steps. 
First, the endpoints of each block are unquantized as follow :
\begin{equation}
    \label{eq:unq} % \unq is defined in utils/command so we can change it ez
    \unq{e}_i = \frac{ a2^{16}e_i + 2^{15} }{2^b}, \quad i = 1,\ldots, 4.
\end{equation}
Where $a = \nicefrac{31}{64}$ for the unsigned BC6 mode, $a=\nicefrac{31}{32}$ for the signed BC6 mode and $b$ is the number of bits used to quantize the endpoints.
This essentially maps the endpoint values into a valid range.
The second step consists in interpolating the endpoints linearly:
\begin{equation}
    \label{eq:interp}
    y = P_k \odot (e_1 + 2^qx(\unq{e}_2 - \unq{e}_1)) + \neg P_k \odot (\unq{e}_3 + 2^qx(\unq{e}_4 - \unq{e}_3)),
\end{equation}
\RV{
where $P_k$ is the binary mask associated with partition $k$ and $q$ is the number of bits used to quantize the pixel indices.
}
This will give a value $y \in [-31743, 31743]$ and whose bits are finally re-intrepreted as a half precision number \cite{ieee}. 
This cast is a non-linear transformation that can be simulated with the following operation
\begin{equation}
    \label{eq:magic1}
    w = 2^{h(y)-14} \left( \frac{y}{1024}-h(y) \right),
\end{equation}
where $h(y) = \max \left( \lfloor \nicefrac{(y-1)}{1024} \rfloor -1,0 \right) $.

Therefore, to simulate the hardware BC6 decompression for each pixel within a block, we unquantize and mix the endpoints using eq. (\ref{eq:unq}) and eq. (\ref{eq:interp}), and finally simulate the bit re-interpretation operation with eq. (\ref{eq:magic1}) resulting in the final value $w$. 
All these operations being almost-everywhere differentiable, it allow us to backpropagate through the BC6 decompression, when the partitions are fixed.

\subsection{Hardware compliant filtering}
\label{sec:bcntfilter}

In a real-time rendering context, sampling a texture at a particular scale is often done via mipmapping.
This technique consists in explicitly storing in memory a version of the texture at different scales, also known as mip, and performing trilinear filtering.
This involves doing a texture lookup and bilinear filtering on the two closest mipmap levels (one higher and one lower), and then linearly interpolating the results. 
In general, this approach stabilizes sampling performance as it fixes the number of processed pixels for each query independently of the scale values.

\begin{figure}
    \centering
    \includegraphics[width=0.8\linewidth]{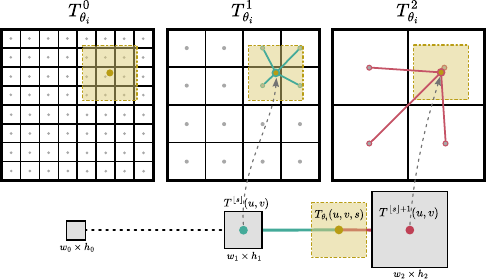}
    \caption{When sampling a feature at a scale level $s$, the block compressed features are filtered using a trilinear interpolation. This requires sampling the two closest mip levels $\lfloor s \rfloor$ and  $\lfloor s \rfloor + 1$ using a bilinear interpolation than linearly mixing the results.}
    \label{fig:bcfiltering}
\end{figure}

Since our goal is to integrate the neural material model in a real-time environment, we propose to rely on trilinear interpolation to filter the neural features $T_{\theta_i}$. 
This makes it possible to exploit hardware accelerated texture filtering to sample the corresponding neural textures during rendering.
To do so, we consider that each feature layer $T_{\theta_i}$ is composed of a pyramid with $S_i$ block-based mips, $T_{\theta_i} =\{T_{\theta_i}^0, \ldots, T_{\theta_i}^{S_i}\}$, of decreasing sizes.
Each block-based mip has independent parameters that will be adjusted during the training process.
In this setting, the scale parameter $s \in [0, S_i]$ refers to the level at which the features $T_{\theta_i}$ are sampled.
This is done as follows:
\begin{equation}
    T_{\theta_i}(u,v,s) = \lambda T^{\lfloor s\rfloor + 1}(u,v) + (1-\lambda)T^{\lfloor s\rfloor}(u,v),
\end{equation} 
where, $\lambda = s - \lfloor s\rfloor$ and $T_{\theta_i}^k(u,v)$ is the bilinear interpolation from the $k^{th}$ block based mip at coordinates $(u,v)$ (fig. \ref{fig:bcfiltering}).

\section{Implementation details}
\label{sec:practical}

In this section, we present all the practical details related to the encoding and decoding of PBR material information with our block based neural features.

\subsection{Material encoding}
\label{sec:material_encoding}

\paragraph*{Model structure}
In our implementation, we use 4 sets of neural block-based features $\{T_{\theta_0}, T_{\theta_1}, T_{\theta_2}, T_{\theta_3}\}$ with size $w_i = h_i = 2^{n_i}$. 
As described in section \ref{sec:bcntfilter}, each $T_{\theta_0}$ consists of a pyramid with several block-based mips of decreasing power of two sizes. 
\RV{
We use a simple MLP with RELU activations as our decoder network and adjust the size of the output according to the endcoded material.
}

\paragraph*{Partition selection strategy}\label{sec:partition_selection}
As mentioned in section \ref{sec:BCforward}, we can backpropagate through the simulated BC6 decompression only if the partition~IDs for each block are fixed.
In order to maximise the reconstruction quality, it is necessary to select the most optimal partition for each block. 
However, choosing the one with the lowest error during training is not straightforward.
\RV{
Indeed, the optimal values of $l_1$, $l_2$ and $x_i$ might vary significantly depending on which partition $k$ is considered. 
}
This means every-time the partition changes, these parameters need to re-adapt to the new set of pixels resulting in training instability. 
To overcome this issue, one approach would be to learn in parallel all the parameters for each possible partition for each block.
This is unpractical and would increase both the memory and time needed to learn the material model.
Instead of that, we propose to consider the partitions as hyper-parameters, and thus fix them before the training.
Since randomly setting their values would possibly lead to non-optimal partition attribution, we therefore propose to learn the material model in two steps. 
First, we perform a quick training using a set of features with unconstrained parameters, i.e., move freely in 3D space and not constrained on a line segment.
Then, we initialize the block-based features by compressing the unconstrained features with the BC6 algorithm. 
This provides us both a partition selection strategy and a relevant initialization of the parameters within each block.
In our experiments, we train the model for 5k iterations with unconstrained parameters, then use the result to initialise the block based features. 
While this does not guarantee the selection of optimal partitions, we found that it improves the reconstruction quality and consistency of the results.

\paragraph*{Reference material sampling}
In our experiments, we consider reference materials with $S$ mipmaps and sample them at random during training by doing a bicubic filtering on the closest two mips then linearly interpolating the results.
More precisely, we process batches of 512x512 uniformly sampled uv-grids, for each batch we also sample a continuous scale parameter $s$ uniformly in $[0, S]$ where $S$ is the number of mipmap.
Both the reference material and the neural material (sec.~\ref{sec:bcntfilter}) are then sampled at $(u,v,s)$ and the loss is computed as the mean squared error between them. 
For more details on the training parameters, we refer to the results in section~\ref{sec:results}.

\subsection{Real-time Decoding}
\label{sec:inference}

We export each trained feature layer as a mipmapped BC6 texture.
\RV{
The block-based structure of these features simplifies this operation as it only involves the quantization and encoding of the endpoints and pixel indices with 6 bits and 3 bits respectively.
}
The network's weights are stored directly in a binary buffer as fp16 values since their size is negligible. 

Multiple methods can be employed to integrate our model into a real-time renderer. 
\RV{
Our implementation is rather straightforward and simply consists in loading the neural textures and weights in GPU memory and proceed as follows (fig \ref{fig:overview}).
}
First, we use the partial derivatives of the input pixel's $uv$ coordinates to compute the sampling scale $s_i$ for each neural texture $T_{\theta_i}$
\begin{equation}
    s_i = \log_2 \left( \max \left( \left| \nicefrac{\partial uv}{\partial u} \right|, \left| \nicefrac{\partial uv}{\partial v} \right| \right) \right) + b_i,
\end{equation} 
where $b_i = \log_2\left(\max(\nicefrac{h_i}{h}, \nicefrac{w_i}{w})\right)$ is a bias value that depends on the neural texture's and render resolutions, respectively $w_i \times h_i $ and $w \times h$. 
Then, we sample each of the neural textures using the GPU's hardware filtering capabilities and pass the resulting values to the decoding shader. 
Within this shader, the sampled values processed through a sequence of functions that perform vector-matrix multiplications with the model's weights and the return the material information. 
When rendering a scene with multiple neural materials, we store all the different networks weights in a single buffer and use a material id value as an offset to access a specific model's weights.
Finally, the pixel is shaded.

\section{Results}
\label{sec:results}

% In this section, we demonstrate the efficiency and performance of our method on various examples.
% We describe our experimental setup in section~\ref{sec:expsetup}, then we showcase the results in section~\ref{sec:quality} and the decoding time performance in section~\ref{sec:performances}.

% \begin{figure}
%     \centering
%     \includegraphics[width=\linewidth]{figures/06-dataset.png}
%     \caption{\RV{Our evaluation dataset includes 12 material texture sets attached with 3D objects. Our implementation can easily render a scene with 144 objects in real-time achieving an average framerate of about 200fps.}}
%     \label{fig:dataset}
% \end{figure}

\subsection{Experimental setup}
\label{sec:expsetup}
\paragraph*{Evaluation dataset}\label{sec:dataset}
\RV{
We gather a dataset of 19 materials from polyhaven.com, 12 of which are accompanied with a 3D model.
All of the material information consist of at least 9 channels that are stored in a set of three textures : albedo, normals, and arm. 
% We gathered a dataset of twelve textured 3D models (fig.~\ref{fig:dataset}) from the public library PolyHaven \cite{polyhaven}. 
% The material information for each object is constituted of eight channels, and stored into three textures : albedo, normals and arm.
% The resolution of each texture is $2048 \times 2048$ (2K).
The albedo layer has 3 channels and contains the diffuse color information.
The normals layer has 3 channels and contains surface normals with respect to a local tangent frame. In practice, our model learns the x and y components since the z value can be reconstructed.
The arm layer has 3 channels and contains the ambient occlusion, surface roughness and metalness parameters.
8 of the materials in the dataset contains an additional displacement channel that is stored in its own texture. 
Note that our model is not restricted to learn only this specific material representation but can handle materials with arbitrary number of layers.
}

\paragraph*{Model configuration.}
In our experiments, our models consisted of four mipmapped square shaped block compressed features.
Table~\ref{tab:model_config} details the resolution of each of the feature layers along with the corresponding number of mips. 
Our decoder network consist of a small \MLP $~$with 12 inputs and one hidden layer of dimension 16.
% A large one with 2 hidden layers of dimension 64. 
% Additionally, we concatenate to the values sampled from the embedding to positional encoding information similar to \cite{ntc2023}. 
% This increases the dimension of the input to 20.  
The output of the network depends on the number of channels in the material texture set.
For the rest of this paper, we will refer to our model by the name of its block compressed features configuration and the size of its network. 
For instance \NBC0.5K refers to the model with a resolution of $T_{\theta_0}$ equal to 512.

\begin{table}[]
\caption{our Neural block-compressed feature configuration.}
\label{tab:model_config}
\resizebox{\columnwidth}{!}{%
\begin{tabular}{@{}ccc|cc|cc|cc@{}}
\toprule
 & \multicolumn{2}{c|}{$T_{\theta_0}$} & \multicolumn{2}{c|}{$T_{\theta_1}$} & \multicolumn{2}{c|}{$T_{\theta_2}$} & \multicolumn{2}{c}{$T_{\theta_3}$} \\ \cmidrule(l){2-9} 
 & \cellcolor{cellcolor}\textit{res} & \textit{mips} & \cellcolor{cellcolor}\textit{res} & \textit{mips} & \cellcolor{cellcolor}\textit{res} & \textit{mips} & \cellcolor{cellcolor}\textit{res} & mips \\\midrule
\multicolumn{1}{l|}{\NBC 0.5k} & \cellcolor{cellcolor}512  & 8  & \cellcolor{cellcolor}256  & 7 & \cellcolor{cellcolor}128 & 6 & \cellcolor{cellcolor}64  & 5 \\
\multicolumn{1}{l|}{\NBC 1k}   & \cellcolor{cellcolor}1024 & 9  & \cellcolor{cellcolor}512  & 8 & \cellcolor{cellcolor}256 & 7 & \cellcolor{cellcolor}128 & 6 \\
\multicolumn{1}{l|}{\NBC 2k}    & \cellcolor{cellcolor}2048 & 10 & \cellcolor{cellcolor}1024 & 9 & \cellcolor{cellcolor}512 & 8 & \cellcolor{cellcolor}256 & 7 \\
\multicolumn{1}{l|}{\NBC 2k++}    & \cellcolor{cellcolor}2048 & 10 & \cellcolor{cellcolor}2048 & 10 & \cellcolor{cellcolor}512 & 8 & \cellcolor{cellcolor}256 & 7 \\ 
\bottomrule
\end{tabular}%
}
\end{table}

\paragraph*{Training parameters.}
We use the Adam stochastic gradient descent algorithm \cite{kingma2014adam} with an exponential decay learning rate scheduler. 
We set a different learning rate for both the block compressed features and the MLP in order to balance the gradient backpropagation.
We train the models for a total of 205k iterations in two phases.
In the first phase we train our model with unconstrained features for 5k iterations using learning rates of $5\times10^{-2}$ and $10^{-3}$ for the features and the MLP respectively, and a decay parameter $\gamma = 0.9995$.
The second phase starts by initializing the block compressed features from unconstrained ones as described in section \ref{sec:material_encoding}. 
Then we train the model for 200k iterations using a learning rate of $10^{-2}$ for the features and $10^{-3}$ for the MLP with decay parameter $\gamma = 0.9999$.
For reference, our \NBC1K requires about 140 minutes to be trained for the 205k iteration on an NVIDIA RTX2070 with our implementation using the PyTorch library \cite{NEURIPS2019_9015}. Note that we observed that the model already outputs decent results after 10k steps, in approximately 15 minutes.

\paragraph*{Compared methods.} 
We compare our method with standard BC compression, ASTC \cite{nystad2012adaptive} and our implementation of Vaidyanathan et al.'s NTC \cite{ntc2023}.
Both BC and ASTC compressed textures were generated using NVIDIA's texture tools exporter \cite{texturetool} with a compression quality set to the highest possible setting.
\RV{
For BC compression, we used the BC5 format to store the normal layer and the BC1 for the rest for rest of the texture set (the albedo, \ARM and displacement textures when available).
}
% This BC configuration is standard and used in most of real-time applications such as video games. 
For ASTC, we used a $12 \times 12$ blocks as it matches the size of our \NBC1K configuration.
In the context of our experiments, where the reference material resolution is $2048\times2048$, the highest resolution of NTC's features grid of set to 512 for NTC0.2 and NTC0.5, and 1024 for NTC1.0. 

\paragraph*{Considered metrics.}
Quantifying the visual quality of an image is still an open problem as no metric can effectively align with human perception.
This is especially the case when the types of distortions introduced by the compression methods are different. 
For instance, traditional BC and ASTC methods tends to have more blocky artefacts while neural methods such as ours and NTC tend to exhibit color shift and feature bleeding. 
For this reason we mainly rely on the PSNR value as it is directly linked to the loss we are optimizing.
In this sense, the PSNR can be seen more as a proxy for the method's capacity to approximate the reference data and not as a measure of visual quality.
\RV{
We also include SSIM \cite{ssim} and FLIP \cite{flip} values as they are supposed to be more in line with human perception.
We compute our metrics for each mip level and aggregate the values as described by Vaidyanathan et al. in \cite{ntc2023}.
}

\begin{table*}[]
\caption{Evaluation of the reconstruction metrics (PSNR, SSIM and FLIP) on the whole dataset. For PSNR, we averaged the mean-squared error over all the mips (as proposed in \cite{ntc2023}) and all the models then computed the PSNR.}
\label{tab:metrics-static}
\resizebox{\linewidth}{!}{%
\begin{tabular}{@{}cclclclcl|cccccc|cc|cc@{}}
\toprule
 &
  \multicolumn{2}{c}{} &
  \multicolumn{2}{c}{} &
  \multicolumn{2}{c}{} &
  \multicolumn{2}{c|}{} &
  \multicolumn{2}{c}{} &
  \multicolumn{2}{c}{} &
  \multicolumn{2}{c|}{} &
  \multicolumn{2}{c|}{\textbf{BC}} &
  \multicolumn{2}{c}{\textbf{ASTC12}} \\
  \multirow{-2}{*}{} &
  \multicolumn{2}{c}{\multirow{-2}{*}{\textbf{\NBC-0.5K}}} &
  \multicolumn{2}{c}{\multirow{-2}{*}{\textbf{\NBC-1K}}} &
  \multicolumn{2}{c}{\multirow{-2}{*}{\textbf{\NBC-2K}}} &
  \multicolumn{2}{c|}{\multirow{-2}{*}{\textbf{\NBC-2K++}}} &
  \multicolumn{2}{c}{\multirow{-2}{*}{\textbf{NTC0.2}}} &
  \multicolumn{2}{c}{\multirow{-2}{*}{\textbf{NTC0.5}}} &
  \multicolumn{2}{c|}{\multirow{-2}{*}{\textbf{NTC1.0}}} &
%   % \cellcolor[HTML]{EFEFEF}\textit{1spp} &
%   \textit{bilinear} &
%   % \cellcolor[HTML]{EFEFEF}\textit{1spp} &
%   \textit{bilinear} &
%   % \cellcolor[HTML]{EFEFEF}\textit{1spp} &
%   \textit{bilinear} &
  \cellcolor[HTML]{EFEFEF}\textit{1024} &
  \textit{2048} &
  \cellcolor[HTML]{EFEFEF}\textit{1024} &
  \textit{2048} \\ 
  \midrule
  \multicolumn{1}{c|}{PSNR $(\uparrow)$} &
  \multicolumn{2}{c}{\RV{28.70}} &
  \multicolumn{2}{c}{\RV{31.83}} &
  \multicolumn{2}{c}{\RV{36.02}} &
  \multicolumn{2}{c|}{\RV{37.86}} &
  % \cellcolor[HTML]{EFEFEF}20.04 &
  \multicolumn{2}{c}{\RV{31.90}} &
  % \cellcolor[HTML]{EFEFEF}30.99 &
  \multicolumn{2}{c}{\RV{33.92}} &
  % \cellcolor[HTML]{EFEFEF}26.98 &
  \multicolumn{2}{c|}{\RV{37.13}} &
  \cellcolor[HTML]{EFEFEF}\RV{28.49} &
  \RV{36.16} &
  \cellcolor[HTML]{EFEFEF}\RV{26.82} &
  \RV{30.48} \\
  \multicolumn{1}{c|}{SSIM $(\uparrow)$} &
  \multicolumn{2}{c}{\RV{0.85}} &
  \multicolumn{2}{c}{\RV{0.90}} &
  \multicolumn{2}{c}{\RV{0.95}} &
  \multicolumn{2}{c|}{\RV{0.96}} &
  % \cellcolor[HTML]{EFEFEF}0.90 &
  \multicolumn{2}{c}{\RV{0.89}} &
  % \cellcolor[HTML]{EFEFEF}0.92 &
  \multicolumn{2}{c}{\RV{0.92}} &
  % \cellcolor[HTML]{EFEFEF}0.95 &
  \multicolumn{2}{c|}{\RV{0.93}} &
  \cellcolor[HTML]{EFEFEF}\RV{0.84} &
  \RV{0.96} &
  \cellcolor[HTML]{EFEFEF}\RV{0.78} &
  \RV{0.89} \\
  \multicolumn{1}{c|}{FLIP $(\downarrow)$} &
  \multicolumn{2}{c}{\RV{0.142}} &
  \multicolumn{2}{c}{\RV{0.099}} &
  \multicolumn{2}{c}{\RV{0.069}} &
  \multicolumn{2}{c|}{\RV{0.059}} &
  % \cellcolor[HTML]{EFEFEF} &
  \multicolumn{2}{c}{\RV{0.089}} &
  % \cellcolor[HTML]{EFEFEF} &
  \multicolumn{2}{c}{\RV{0.069}} &
  % \cellcolor[HTML]{EFEFEF} &
  \multicolumn{2}{c|}{\RV{0.057}} &
  \cellcolor[HTML]{EFEFEF} \RV{0.099} &
   \RV{0.036} &
  \cellcolor[HTML]{EFEFEF} \RV{0.143} &
  \RV{0.082} \\ \midrule
  \multicolumn{1}{c|}{size (MB)} &
  \multicolumn{2}{c}{0.44} &
  \multicolumn{2}{c}{1.77} &
  \multicolumn{2}{c}{7.08} &
  \multicolumn{2}{c|}{11.08} &
  \multicolumn{2}{c}{0.93} &
  \multicolumn{2}{c}{2.26} &
  \multicolumn{2}{c|}{4.53} &
  \cellcolor[HTML]{EFEFEF}2.91 &
  11.65 &
  \cellcolor[HTML]{EFEFEF}0.53 &
  2.11 \\
  \bottomrule
\end{tabular}%
}
\end{table*}

\begin{figure}
    \centering
    \includegraphics[width=0.8\linewidth]{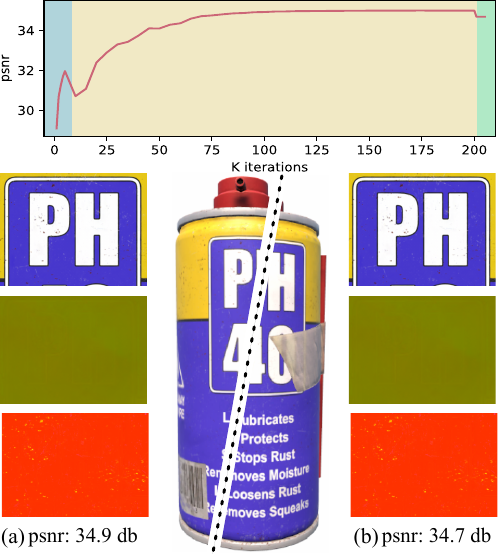}
    \caption{
    Our training process comprises three stages: A warmup stage (blue) with unconstrained neural features. The main training stage (yellow) with Block-based features initialised from the result of the previous stage. Finally, a finetune stage (green) the BC features are quantized.
    The results at the end of the second stage (a) and third stage (b) are visually identical. 
    % Effects of BC compression of the neural features on the decoded textures. The decoded textures before BC compression of the neural features (a) are not visually different from the decoded textures after BC compression (b). The render on the right shows that no artifacts are visible on the final shaded texture.
    }
    \label{fig:bcnt}
\end{figure}

\subsection{Reconstruction quality}
\label{sec:quality}

% In this section, we showcase the capacity of our model to reconstruct the learned PBR material information.
% In a 3D setting, the material texture set are sampled at $uv$ coordinates that does not necesserilly allign with the , due to the 3D object's orientation and distance.
% We emulate this in our experiments by randomly sampling a point within each pixel and comparing our results with a filtered version of the reference data.
% In practice, this is done by adding a random noise to a regular grid of $uv$ points such that their coordinates remains within the pixel's boundary.

\paragraph*{Block-compressed features.} 
\RV{
Figure~\ref{fig:bcnt} shows the evolution of the PSNR value throughout the entire training. 
We start by training a set of unconstrained embedding and use them to initialise the BC features at the 5000th iteration as stated in section \ref{sec:partition_selection}. 
%It is normal that the psnr value drops here since the input to the network suddenly changes due to switching up the embeddings. We train the model for another 2000 iterations before quantizing the trained end-points and the indices. At this stage we freeze the values of the BC features and continue the training for another 1000 iteration for fine-tune the network. While the quantization of the trained features does reduce PSNR value, its impact visually is negligeable.
It is expected that the PSNR value drops at this point since the network's input undergoes a change due to the alteration of the embeddings. The BC constrained model is trained for 200K iterations, after which we quantize the trained end-points and indices. Following this, we freeze the values of the BC features and proceed with training for 1000 more iterations to fine-tune the network. 
Although quantizing the trained features leads to a slight reduction in PSNR value, its visual impact is negligible.  
The material information reconstruction from the raw unquantized fp16 block compressed features (fig.~\ref{fig:bcnt} (a)) and from the exported BC6 block features (fig.~\ref{fig:bcnt} (b)) are visually indistinguishable.
}

% \begin{figure}
%     \centering
%     % \includegraphics[width=\linewidth]{figures/06-gridsize.png}
%     \includegraphics[width=\linewidth]{06-gridsize.pdf}
%     \caption{Effect of the resolution of the neural compressed features on the visual reconstruction. Note that this resolution parameter is directly linked to the compression ratio our method achieve.}
%     \label{fig:gridsize}
% \end{figure}

\begin{figure}
    \centering
    \includegraphics[width=0.8\linewidth]{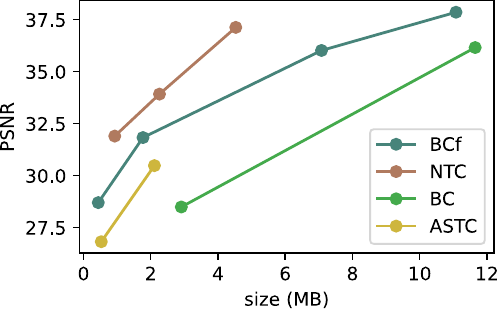}
    \caption{Reconstruction PSNR versus size in megabytes (MB) for all the methods presented in table~\ref{tab:metrics-static}.}
    \label{fig:graph-psnr}
\end{figure}

\begin{figure*}
    \centering
    \includegraphics[width=0.95\linewidth]{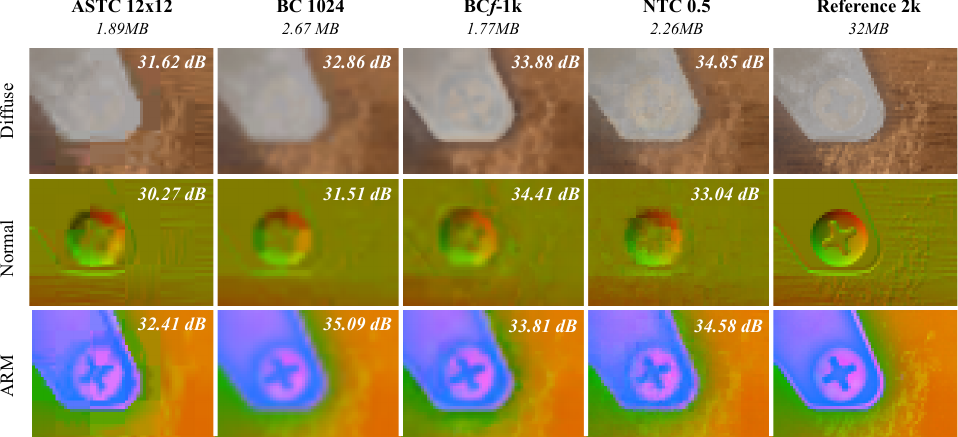}
    \caption{Close-up crop of the Ukulele's material texture set. Comparison between ASTC, BC, our \NBC1K, NTC0.5 methods and the reference 2K material.}
    \label{fig:comparison}
\end{figure*}

\paragraph*{Visual Performance.}
We trained all our model configurations against a reference 2K material and compared them to the methods mentioned in section~\ref{sec:expsetup}. 
The results are gathered in table~\ref{tab:metrics-static} and illustrated in figure~\ref{fig:graph-psnr}. 
In these experiments, the texture set is reconstructed using a regular $uv$ grid aligning with the center of pixels and the metrics are evaluated by comparing the result with the reference 2K material.
Our model outperforms standard BC and ASTC textures as it does not exhibit blocky artifacts and is capable of reconstructing sharper visuals with less memory (see fig.~\ref{fig:comparison}).
It is not surprising to see that the resolution of the neural features has a direct impact on the capacity of the model to reconstruct the material.
\RV{
The higher the resolution, the better the reconstruction. 
However, our model's performance does not scale equally with the increase in resolution and can output a slightly smoother result when compared with NTC \cite{ntc2023} (see fig.~\ref{fig:comparison}).
This is due to our minimalist neural architecture that relies on a very small MLP with one hidden layer of size 16 and an input of size 12 to reconstruct a very complex signal which is particularly well suited for real-time performance. 
While naively increasing the size of the hidden layers does improve the quality, it causes the inference time to increase in a quadratic manner. 
We refer to the supplementary material for additional experiments with larger networks and other configurations.   
}
% Compared to standard BC and ASTC textures our model yields sharper materials and does not exhibit blocky artifacts (see fig.~\ref{fig:comparison}).
% In general, our model reconstructs a slightly smoother material compared to NTC.

\begin{figure}
    \centering
    \includegraphics[width=0.8\linewidth]{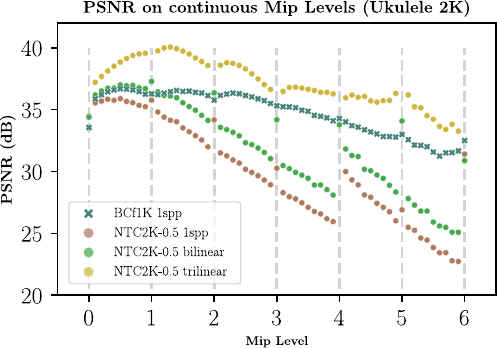}
    \caption{Evolution of the PSNR reconstruction metric over the scales on the Ukulele's material.}
    \label{fig:learnfilter}
\end{figure}

\begin{figure*}
    \centering
    \includegraphics[width=0.95\linewidth]{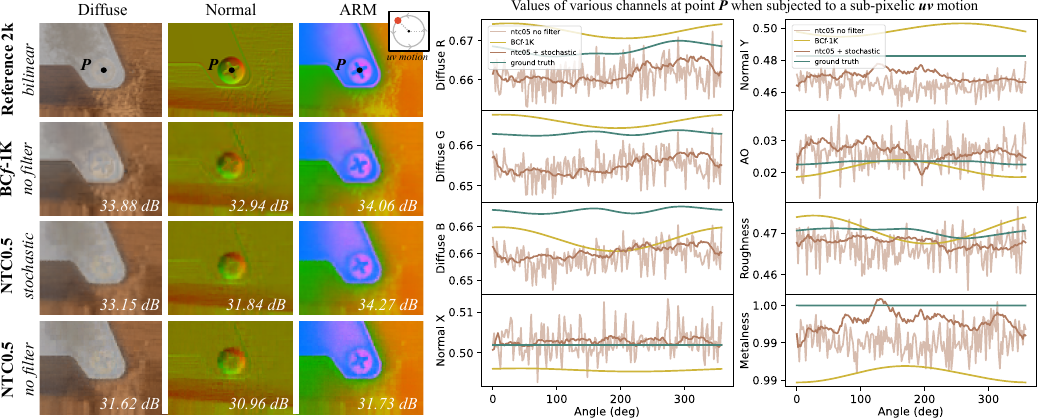}
    \caption{\RV{Close-up crop of the reconstruction of the Ukulele's material when subject to a circular motion. We also track the values of each channel throughout time at point $P$. NTC is not stable when it is evaluated center of the pixel and requires some filtering which smoothen the result. Our approach does not require any filtering and output values that are stable temporally.}
    ~\label{fig:temporal-value}}
\end{figure*}

\begin{figure}
    \centering
    \includegraphics[width=0.95\linewidth]{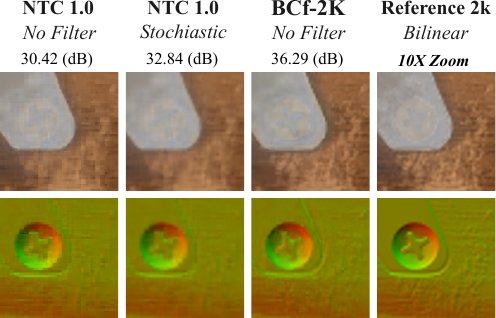}
    \caption{\RV{Reconstuction the Ukulele's texture set with $10\times$ magnification. Without any filtering NTC exhibit blocky artifacts and high frequency noise. Pairing it with a stochiastic filter will yield a smoother result. Our model can handle this without any filtering.}}
    \label{fig:mag}
\end{figure}

\RV{
The main advantage of our approach with respect to NTC, is its capacity to output filtered material information. 
This is particularly important since, in a 3D environment, material texture sets are almost never perfectly aligned with the screen's pixel.
The misalignment between the 3D object and the viewport, along with changes in distance, results in the material being sampled at random points within the $uv$ domain and across continuous scales.  
NTC cannot handle this in a straightforward manner since it is trained by considering a fixed uv grid and lod values.
For instance, in fig. ~\ref{fig:learnfilter}, we compare our method to NTC by evaluating the result while constantly changing the lod value. 
It shows that NTC's performance is not stable when opting to reconstruct the material with 1 sample per pixel (spp) and requires to be paired with a filtering operator. 
In this case, a bilinear or trilinear filtering is not suited for real-time rendering as it requires decoding multiple samples per pixel. 
While temporal stochastic filtering can mitigate this issue, its reconstruction quality is dependant on several factors (jittering pattern, noise distribution, accumulation policy, etc) and does result in a smoother reconstruction in motion (fig. \ref{fig:temporal-value}).
Our approach, on the other hand does not require any filtering and is capable to reconstruct the material texture with 1spp. 
Moreover, the values outputted by our decoder remains stable across time (fig. \ref{fig:temporal-value}) and mip levels (fig. ~\ref{fig:learnfilter}) and can even handle extreme magnifications (fig. \ref{fig:mag}) .
This is not surprising since our model is trained to emulate a continuous filtering operation. 
In this case our network plays the role of a decoder and a filter at the same time.
% Its performance drops significantly when reconstructing the material with 1 sample per pixel (spp) (fig.~\ref{fig:graph-psnr}).
% Figure~\ref{fig:learnfilter} shows that its performance is not consistent across mip level.
% To handle this, NTC requires to follow up the decoding step with a filtering operation which increases the number of network evaluation per pixel.
% For instance, it is required to decode the four neighbour pixel for bilinear filtering or the eight neighbour pixels for trilinear filtering. 
% Our model does not suffer for this issue and is capable of reconstructing a filtered image directly form a random $uv$ value. 
% This makes it more suitable for real-time rendering as only one network evaluation per pixel is necessary. 
}

\subsection{Decoding performance}\label{sec:performances}
% We created a scene consisting of 144 3D objects whose materials are encoded using our neural model. 
% It can be rendered at more than 200 frames per second (fps) using an NVIDIA RTX2070 GPU (fig.~\ref{fig:dataset}).
In order to better illustrate the computational overhead of our neural material model, we rendered a textured full-screen quad at $1080$p, $1440$p  and $2160$p (4K).
We then compared the total rendering time of our \NBC-1K with rendering times obtained using the same setup but with standard BC textures.
The results are presented in table~\ref{tab:performance}.
They clearly show that our method is well suited for real-time environments as the computational overhead on a 4K resolution is only 0.6ms for a full 4K screen.
The small computational overhead is a direct consequence of our block compressed neural features. 
Exporting these features as BC6 textures makes their memory footprint small which makes it possible to increase their resolution.
This allows us to rely on a very small and fast decoder network to reconstruct the material.
More importantly, it enables the use of hardware texture filtering operations to sample the features at render-time which minimizes the overhead.

\begin{table}
\caption{Average time to render a textured full-screen quad using an NVIDIA RTX2070.}
\label{tab:performance}
\centering
\begin{tabular}{@{}cc|cc@{}}
\toprule
 & & \multicolumn{2}{c}{\NBC1K} \\ 
 & \multirow{-2}{*}{BC Textures} & \cellcolor[HTML]{EFEFEF}\textit{total time} & \textit{neural overhead} \\ \midrule
\multicolumn{1}{c|}{$1080$p} & 0.376 ms & \cellcolor[HTML]{EFEFEF}0.555 ms   & 0.179 ms  \\
\multicolumn{1}{c|}{$1440$p} & 0.614 ms  & \cellcolor[HTML]{EFEFEF}0.924 ms   & 0.310 ms  \\
\multicolumn{1}{c|}{$2160$p} & 1.684 ms  & \cellcolor[HTML]{EFEFEF}2.324 ms   & 0.640 ms  \\ \bottomrule
\end{tabular}
\end{table}

% \section{Discussion}
% \label{sec:limitations}

\subsection{Quantization vs. Compression}
\begin{figure}
    \centering
    \includegraphics[width=0.8\linewidth]{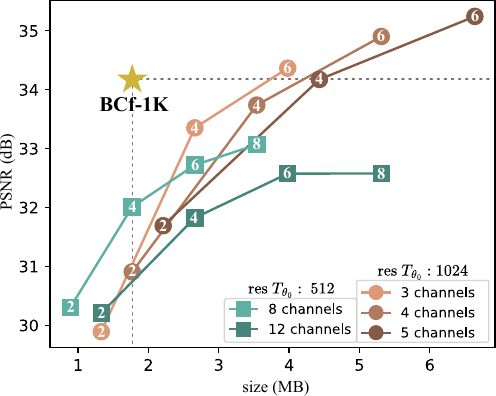}
    \caption{The reconstruction performance of the Ukulele's material using neural features with various number of channels and quantization rate versus Block Compressed ones. The quantization rate is written in white.}
    \label{fig:bitrate-quantization}
\end{figure}

\begin{figure}
    \centering
    \includegraphics[width=\linewidth]{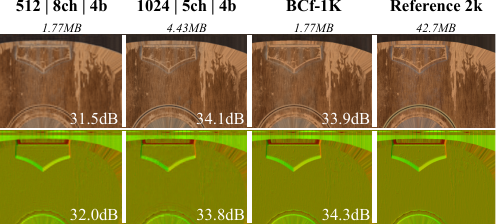}
    \caption{\RV{Texture set reconstruction obtained with neural features of various configurations. Our \NBC1K configuration matches up visually with neural features of dimension 5 and quantized on 4 bits while being 40$\%$ smaller in size.}}
    \label{fig:bitrate-quantization-crop}
\end{figure}

\RV{
As mentioned in section \ref{sec:framework}, there are two possible strategies that can be adopted when it comes to storing the learned neural features with fewer bits. 
We can either quantize their values or maintain them at a high bitrate and compress them. 
Both approaches are not straightforward to implement.  
% Storing the learned neural features efficiently boils down to using fewer bits by either (1) quantizing or (2) compressing them.  
Simply reducing the number of bits downgrades quality. 
This can be mitigated by increasing features's dimension but requires reducing the resolution to maintain manageable memory usage.
For instance, it is possible to match the size of our \NBC1K profile with features of 4 channels quantized with two bits, or by dividing their resolution by two and increasing the number of channels to 8 and quantizing with 4 bits. 
Figure \ref{fig:bitrate-quantization} plots the PSNR values of these configuration for the ukulele model and shows that, when pairing the features with a small network, it is more beneficial to have high bitrate features that are compressed than high dimensional low bitrate values. 
On average, at equal resolution, neural features with 3 channels stored according to be BC6H format performs similarly to ones with 5 channels and quantized with 4 bits resulting in a memory reduction of $40\%$ (fig. \ref{fig:bitrate-quantization-crop}).
Increasing the number of channels and increasing the resolution without increasing the complexity of the network is not a viable strategy.
The resolution of the features has a major impact on the reconstruction quality that cannot be compensated by increasing the number of channels and more importantly the larger input will overwhelm the small network and impact the result in a negative way. 
A larger network and more complex architecture are, thus, needed to fully take advantage of the higher number of input features and compensate for the lower resolution and bitrate as demonstrated by Vaidyanathan et al. \cite{ntc2023}. 
}
% Using the BC formats to store the features limits the number of channels which makes them less efficient when paired with larger network. 
% This in turn requires a larger MLP which is expensive to evaluate. This approach is taken by [VSW23]. NTC0.5 has neural features of dimension 12 and 20 and resolution 0.25xMip0 and 0.125xMip0 at 4bit. 
% We adopt the second approach.
% By compressing the features, they remain encoded a high-bitrate which allows us to have shallow neural features at high-resolution.
% This makes it possible to use a small network at the cost of decompressing the features before being passed to the network. The challenge here is minimizing the decompressions’ computational impact and making the decoder robust to its artifacts. We mitigate these issues by encoding the neural features directly in BC6 format. 
% Our architecture uses 4 sets of FP16 features of dimension three. 
% The impact of the network size on the rendering is considerable. 
% Pairing a small network with NTC’s neural features yields lower quality reconstruction. 
% We observed that to match our quality with small network using the bit reduction approach requires more memory (the resolutions of G0/G1 should be doubled).  
% Consequently, encoding features with BC6 is more memory efficient than a simple bit reduction when a small network is required.

\subsection{Impact of BC partition selection}
\begin{figure}
    \centering
    \includegraphics[width=0.8\linewidth]{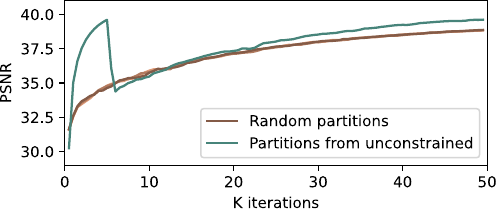}
    \caption{Initializing the partitions of the neural BC feature from a set of unconstrained features yields better result in the long run. After 50k iterations, our initialisation heuristic is ahead by 1 PSNR unit compared to a randomized initialisation.}
    \label{fig:optim_part}
\end{figure}

\RV{
By design, the introduced BC6 constraints restrain the model capability compared to unconstrained features. 
The more the partition shape is aligned with the content of the data in a $4\times4$ block, the better the compression. 
Thus randomly selecting the partitions or forcing a static one is not optimal and will lead to lower reconstruction performance. 
Figure \ref{fig:optim_part} shows that by initializing the partitions from pre-trained unconstrained neural features, as discussed in section \ref{sec:partition_selection}, yields better result in the long run. 
In general, the warmup stage is not required to be long. 
We found that usually 5K iterations are sufficient for the shape of the unconstrained features to stabilise resulting in more suitable partitions. 
On our hardware, this takes up to 3 minutes.
% However, this can be mitigated with a wise choice of BC partition as presented in section \ref{sec:partition_selection}. Figure~\ref{fig:bcpartition} shows the reconstruction quality evolution according to training steps for 1k, 2k and 5k first phase steps and both random and proposed strategy for partition selection. This illustrate the impact of partition selection, and the importance of reaching a plateau for first phase training before fixing the partitions. Indeed, our selection strategy always overcome the random strategy and tends to reach the same asymptotic behaviour than the unconstrained model.
}

\section{Limitations and Future Work}
\label{sec:limitations}
In this section we discuss some of the limitations of our neural material model and offer avenues for future work.

\paragraph*{Reconstruction artifacts} 
Our reconstructed material can experience layer bleeding and color shift.  
Since our model is exploiting correlation between the channels and projecting the data into a smaller dimensional space, we believe these artifacts occur when there is small correlation or their absolute values are too divergent. 
In addition, the use of a decoding network consisting of only one hidden layer makes our model struggle with reconstructing and separating the data from the material layer. 
Even increasing the network size is hardly an option considering the potential overhead. It should be possible to overcome these issues by designing a decoder network that separates the reconstruction of uncorrelated layers. 
In addition to that, we could learn a tone mapping operator to map each output to the corresponding values.
The challenge here, is to keep the decoding network small in order to maintain real-time performance, so we leave this to future work.

\paragraph*{Higher quality reconstruction}
\RV{Depending on the profile, our model may struggle to learn high-frequency details which results in a slightly over-smooth reconstruction.
% This as a consequence of the features being of lower resolutions and lower dimension than that of the reference material.
% In this setting the decoder network must perform some up-scaling in order to compensate. 
This is a direct consequence of having neural features with only 3 channels and a very small decoder network, which prevent the effective use of positional encoding that could have contributed additional high frequency information as shown in \cite{ntc2023}.  
Increasing the number of channels is key here.
However, it is not straightforward in the context our BC neural features as the BC6H format is only capable of compressing RGB images. 
One way to overcome this is to group the high dimensional features into sets of 3 and process them independently. 
In this context, our \NBC2k++ configuration can be seen as one where we have three neural layers where the first one has 6 channels.}
%We leave this research direction to future work.  
%We also want to investigate the usage of non square features (i.e., larger resolution in one dimension). 
%This would allow to encode higher frequencies separately in horizontal and vertical directions while maintaining the same memory usage.
%}\todo{par sur qu'on laisse ce truc d'embeddgin pas carrés}

\paragraph*{Faster material encoding}
In this current version, our PyTorch implementation of the training pipeline requires about 140 minutes for 200K iteration steps. 
For now, we mainly focused on designing a model with a fast decoder which is critical for real-time applications such as video games. 
Reducing the training time is equally important as it is key in making such methods less computationally intensive and more practical in dynamic environments where the assets keep changing. 
There are several avenues that we can employ to improve on this. 
For instance, it is possible to train a generic encoder to initialize the embedding from the reference material in order to have a better starting point and train for fewer iterations as in \cite{zeltner2023real}.
Additionally, we could improve the sampling strategy during training and sample only where it matters. This could also have an impact on the visual quality as the network will focus on minimizing the errors only where it matters.

\paragraph*{Simulating more complex filtering}
Even with simple MLP decoders, our architecture is able to emulate simple texture filtering operations such as bilinear and bicubic filtering. 
We believe that this idea can be explored further by training models to emulate more elaborate filtering operations.
\RV{
For instance it would be possible to sparsely sample the neural textures in an anisotropic manner, i.e., along the major ellipse axis, and train them against a high density anisotropic filtered ground truth. 
This would make it possible to emulate high quality anisotropics filtering with few samples.
}
% One way to improve the filtering is to take into account anisotropy in the scale parameter. 
% In a real-time rendering context, anisotropic filtering is done by generating non-square mipmap in each dimension. These directional mips are then sampled depending on the directional scale parameters $\nicefrac{\partial uv}{\partial u}$ and $\nicefrac{\partial uv}{\partial v}$. 
% We plan to adapt our framework to simulate anisotropic filtering by adapting the architecture of our block-compressed feature layers. 
% Another option we plan to explore, is the capability of learning more complex filters such that the magic kernel \cite{magic2021}.

\section{Conclusion}
\label{sec:conclusion}

This work introduces a novel block-compressed feature layer along with a continuous neural material encoding framework. 
This allows us to design an encoding and a decoding method that fits well within the real-time rendering pipeline constraints. 
We demonstrate the ability of our method to compress PBR materials efficiently and decompress them in real-time in a shader on consumer-level hardware. 
By taking into account memory and time constraints, and by taking advantage of existing hardware operations, we make neural textures usable at large scale in a real-time rendering pipeline, such as a video game engine. 
We hope that our work could serve as a starting point for future work and open the door for more complex use cases, such as learning directional materials or more complex filtering.

\section*{Acknowledgements}
We thank Dr. Heqi Lu for all the discussions and advice and Arnaud Schoentgen for proofreading the paper. 
We also thank all the anonymous reviewers for their insightful comments.  

\bibliographystyle{eg-alpha-doi}
\bibliography{bibliography}              
%==============================================================%

% \input{sections/supplementary.tex}
% \bibliographystyle{eg-alpha-doi}
% \bibliography{bibliography}   

\end{document}